# How LLMs Might Think

Dept. of Philosophy, Texas Tech University, **Joseph Gottlieb** (0001-9014-8487)
Dept. of Philosophy, New York University, **Ethan Kemp** (0009-0004-5871-373X)
Dept. of History and Philosophy of Science, University of Pittsburgh, **Matthew Trager** (0009-0001-1204-6378)

**Abstract**

Do large language models ("LLMs") think? Daniel Stoljar and Zhihe Vincent Zhang have recently developed an *argument from rationality* for the claim that LLMs do not think. We contend, however, that the argument from rationality not only falters, but leaves open an intriguing possibility: that LLMs engage only in arational, associative forms of thinking, and have purely associative minds. Our positive claim is that if LLMs think at all, they likely think precisely in this manner.

## 1. Introduction

Turing (1950) asked if machines can think. We suspect the answer is yes, at least in principle. But our question is more specific: do large language models ("LLMs") think? Mind space—the space of possible minds—is large. So, if LLMs *do* think, they may not think like we do.

Thinking involves not just having individual thoughts. It involves *mental transitions*, or *movements between* thoughts ("trains of thought"). We understand "thoughts" broadly to include both occurrent assertoric thoughts (or judgements) and standing beliefs and desires. When we ask whether LLMs think, we are interested in LLMs broadly within the current paradigm; roughly, artificial neural networks with a transformer-based architecture, trained via stochastic gradient descent, and fine-tuned via processes like reinforcement learning through human feedback ("RLHF"). We also assume that thinking can be entirely nonconscious (Berger, 2014).

Stoljar and Zhang (2024) argue that LLMs do not think. Their *argument from rationality* goes like this:

R1. If LLMs think, they are rational.

R2. If LLMs think, they are not rational.

∴ LLMs do not think.

We have two aims. After some further table-setting in Section 2, our first aim, in Section 3, is to show that this argument is unsound: R1 is false. But this leaves open an intriguing possibility.

Consider that the consequent of R2 might be interpreted in two ways:

**IR**: LLMs are *ir*rational.

**AR**: LLMs are *a*rational.

By our lights, if IR were true, LLMs are thinkers. So, what is true is not R1, but this conditional: if LLMs are irrational, they *do* think. Yet what of AR? If LLMs are arational, are they incapable of thinking? Not necessarily. Arationality is only consistent with non-thinking. LLMs might be arational in the way rocks are arational (by not thinking at all), but they might be arational in some other way. Perhaps within the space of mind design there are purely arational thinkers. After all, thinking is not "one kind of thing" (Quilty-Dunn & Mandelbaum, 2019, p. 151). There may be other, arational forms of thinking.

And that's what we suspect LLMs are up to, if anything. So, our second aim, in Section 4, is to answer this question: if LLMs think, do they think in rationally assessable ways—in other words, do they think inferentially ("i-thinking")? Or do they think in arational ways—in other words, do they think associatively ("a-thinking")? Assuming LLMs have mental states at all, we claim that they are arational minds, and traffic only in associative mental transitions. That is: if LLMs think, they only a-think.

## 2. A Word About Methodology

As a first pass, to make an inference is to engage in a certain kind of "movement in thought" (Boghossian, 2018, p. 55)—paradigmatically, moving from one belief (e.g., *that it rained last night*) to another (e.g., *that the ground is wet*), where there (typically) is a transfer of epistemic warrant between them.[1] How could we tell whether LLMs i-think in (something like) this way, or engage in any other mental activity? Questions of method are invariably controversial here. And unsurprisingly so, given that we are still in the early days of investigating modern LLMs.[2] Nonetheless, it will help to lay down our methodological cards. While we will be relatively brief for now, we will have cause to return to methodological matters later in Section 4.1.

A common investigative option is a *black-box* strategy. An example of this is Bubeck et al.'s (2023) work arguing that GPT-4 shows "real sparks" of general intelligence (Grzankowski, 2024). On this approach, a system is interpreted solely in terms of its inputs and outputs; what goes on between them is left opaque. Still, although the system is a "black box", robust behavioral observations are deemed sufficient to justify psychological attributions. A bit crudely: GPT-4 *looks* intelligent, so it is.

This strategy may seem natural for LLMs, given their scale and complexity (even GPT-4 was rumored to have 1.7 trillion parameters). But some researchers are optimistic that they can profitably break into the black box and learn what the weights represent (if anything) and how exactly processes like gradient descent land on solutions. This young project of *inner interpretability* can proceed via various methods (e.g., probing, causal intervention), but what is common amongst them is the assumption that the inner mechanisms of LLM can be uncovered.[3]

---

[1] Although "inference" is used as a technical term for the process via which LLMs generate new text, whether this *really is* a form of inference (and so a form of thinking) is part of our question.
[2] Foundational research in natural language processing goes back to at least the 1960s with ELIZA, but the advent of the germane kind of LLMs roughly coincides with Google's "Attention Is All You Need" (2017).
[3] For probing, see, for example, Belinkov (2021) and Li et al. (2022). For the limitations of probing, and the corresponding benefits of causal intervention methods, see Milliere and Buckner (2024).

In a way, inner interpretation has as its remit something outside our main focus. This is because our thesis concerns the *kinds of transitions* LLMs make between their representations (assuming they represent at all), not *what* they are representing. However, as we will see, there are some views of inference that put constraints on the kind of representational *vehicles* inferences operate over. So conditional on those theories, interpretability results may be germane. This will prove to mostly help our thesis rather than hurt it.[4]

The crux of our approach though, will look not to interpretability, but a particular operationalization of the distinction between associative and inferential mental transitions: one where associative transitions are only modulable by certain non-rational interventions. And we will claim that the only ways to modulate LLM mental transitions (assuming they have them) are ways that one would modulate associations, not inferences. To operationalize in this way is to focus on observable behaviors, and to that extent, bears some resemblance to black-box approaches. But we certainly do not see *operationalism* as a replacement for underlying metaphysics (cf. Quilty-Dunn & Mandelbaum, 2019, p.160), and in any case, the precise contours of this-or-that methodology is not an issue to litigate here. Whatever is offered as a test for LLM thinking, it cannot rest on the ephemeral. Our approach honors that constraint.[5]

---

[4] By "interpretability", we mean something like what Chalmers (2025) calls *representational interpretability*. Chalmers notes that there are different uses of "interpretability" in AI research, including a general usage where the term picks out any activity meant to render AI systems intelligible. Within representational interpretability, there is, as Chalmers points out, also *propositional interpretability* (understanding the propositional attitudes an AI system is using) and *conceptual interpretability* (understanding the concepts an AI system is using). While our thesis is consistent with the viability of propositional interpretability, it is more at home with conceptual interpretability (since one might think of associations as holding between concepts).

[5] Another strategy we will not entertain argues that because LLMs play chess (write code, give recipes for a soufflé, or design an exercise program), they make inferences. We do not wholesale reject the viability of such a "capabilities approach". However, here is something to be aware of: that a subject S engages in some capacity C (or cluster of capacities), and C involves inference, doesn't entail that S *itself* makes inferences when exercising C. *Seeing* is a capacity that involves inference (to solve underdetermination problems regarding distal conditions), but no one thinks that *we* perform inferences when we see. Inference here is *sub-personal*, done by our early visual systems (Drayson, 2014). Thus, even if LLMs have a capacity that requires inference, this does not mean that *person-level* mental transitions of LLMs made in service of exercising that capacity

## 3. The Argument from Irrationality

Recall R1 from the argument from rationality:

R1. If LLMs think, they are rational.

Following Stoljar and Zhang, let a subject S be *rational* if and only if, in most cases, S performs correct inferences. Let a *p-belief* be a premise belief, and a *c-belief* be a conclusion belief. Stoljar and Zhang (2024, pp. 4-5) assume that *S* performs a *correct* inference if and only if S's p-belief(s) confirms or entails their c-beliefs. The idea underlying the consequent of R1 for Stoljar and Zhang, then, is that LLMs' inferences, if they made them, would be incorrect, and thus *irrational*: their p-beliefs do not confirm or entail their c-beliefs (Stoljar & Zhang, 2024, p. 6).[6]

Stoljar and Zhang tell us that R1 is not an analytic or metaphysically necessary truth, and that it is a "by-and-large" claim, consistent with well-known results coming out of the heuristics and biases literature (e.g., Kahneman & Tversky, 1974). However, these are not reasons to affirm R1; they are only reasons not to deny it. In favor of R1, Stoljar and Zhang only say this:

> Suppose you have a thinker who in no case at all responds correctly to its evidence; in no case at all, that is, can we identify a [p-belief] that entails or confirms any [c-belief] In such a case, it would be better to assume that the individual in question isn't a thinker in the first place rather than to assume that

---

are themselves inferential. Our hope is that the present operationalization is better suited at uncovering what *LLMs* do, not what a sub-component of them does. Granted, in emphasizing that our concern is what LLMs do, one might press further and ask whether we mean LLMs *qua* the underlying model (e.g., GPT-5) or the very many *instances* of this model (e.g., implementations of GPT-5). We will often speak in terms of the former, but we hope nothing turns on this very complicated issue.

[6] If an LLMs' p-beliefs failed to confirm its c-beliefs and it were irrational, Stoljar and Zhang contend this would be because the LLM was guilty of rife use-mention fallacies: its p-beliefs are always about statistical patterns of tokens of the English language, not about the world. Stoljar and Zhang concede that LLM's c-beliefs might *also* be about words or probabilities of token generation but see this as cold comfort: LLMs would make correct inferences, but only about tokens (at best).

> they are a thinker who is in this egregious sense irrational. It is considerations of this sort that allow us to arrive at [R1]: if [the LLM] thinks, it is rational (Stoljar & Zhang, 2024, pp. 12-13).[7]

We are doubtful. Why "would [it] be better to assume that the individual in question isn't a thinker", rather than a *thinker* who is "in this egregious sense irrational"?

Here is a more precise rendition of our claim:

**Irrational-Think**: If S is being irrational from $t_1$ to $t_2$, S thinks from $t_1$ to $t_2$.[8]

Our argument for Irrational-Think goes like this:

I1. If S is being irrational from $t_1$ to $t_2$, then S is making incorrect inferences from $t_1$ to $t_2$.

I2. If S is making incorrect inferences from $t_1$ to $t_2$, S is making inferences from $t_1$ to t2.

I3. If S is making inferences from t1 to $t_2$, S is thinking from $t_1$ to $t_2$.

∴ If S is being irrational from $t_1$ to $t_2$, S is thinking from $t_1$ to $t_2$. [Irrational-Think]

Paralleling the argument from rationality, call this the *argument from irrationality*.

Premises I1 and I3 are innocuous. As for I1, Stoljar and Zhang (2024, p. 4) endorse it. And for good reason. If S is irrational, it is because it does something irrational. In the relevant sense of "rational", the only thing it could do which merits describing S as irrational is making incorrect inferences. Stoljar and Zhang (2024, p. 5) also endorse I3, and again for good reason. Inferring is a form of thinking. This leaves I2.

Here is an initial point in favor of I2: the phrase "incorrect inference" is unlike "kosher bacon". One cannot infer "that's a piece of bacon" from "that's a piece of kosher bacon." By contrast, it seems that one can infer "John made an inference" from "John made an incorrect inference." Imagine a math teacher grading an assignment. Upon noticing that their student performs an incorrect inference, the teacher would not conclude that the student fails to make an inference *at all*, but that the *inference* simply does not meet some standard of correctness.

[7] As for other proponents of R1, Stoljar and Zhang mention Camp (2009, 286).

[8] Irrational-Think is indexed to times because thinking is a process, and processes unfold over time. If S is irrational *simpliciter*, we can adjust the consequent of Irrational-Think to say that S *can* think. We return to this modal version of Irrational-Think shortly.

While theories of inference are legion (e.g., Harman, 1986; Kornblith, 2012; Quilty-Dunn & Mandelbaum, 2018; Boghossian, 2014; Buckner, 2019), they all converge on I2, or something near enough. To keep things manageable, we limit our discussion to three theories: Boghossian's (2014) *taking account*, Quilty-Dunn and Mandelbaum's (2018) *bare inferential transitions*, and Buckner's (2019) *categorization account*.

On the *taking account*, when a subject S infers c-beliefs from p-beliefs, S must *take* his p-beliefs to support his c-beliefs (Boghossian, 2014, p. 5). Exactly what *taking* involves may seem like an especially pressing question in the context of LLMs, given its ring of psychological sophistication, although Boghossian (2014, p. 6) denies that it requires meta-representation. This issue, however, need not detain us.[9] More immediately pressing is that *taking* is wrapped up with the fact that inference has a quality dimension: we speak of whether your p-beliefs provides good reasons for your c-belief. Taking is something *we* do and so tracks the sense in which we can be held responsible for reasoning poorly (Boghossian, 2018, pp. 59-60).

This suggests I2. Suppose S comes to believe that *q* because she believes that *p* and believes or "intellectually intuits" that *p* supports *q* (Boghossian, 2018). Yet suppose also that *p* is *not* in fact a good reason upon which to base one's belief that *q*. This is *bad taking*. Is bad taking an instance of taking? It seems so. After all, the very same internal mental process is occurring. As such, incorrect inferences are inferences. Consider an analogy with perception. Suppose you see a stick as bent, but it is actually straight. This is a visual illusion, but it is still a case of genuine perception (i.e., a case of genuine seeing). True, you did not get all the stick's properties right, but you did not need to.

[9] To note, we are not here presupposing any theory of inference, and we are not yet making any claims concerning whether *LLMs* in fact i-think. At this point, we are just concerned with inference itself. (And our positive conditional thesis to come only turns on an operationalization of the a-thinking versus i-thinking distinction.)

Illusory perception is not a failure to perceive. Rather, it is perceiving minus some property (e.g., *veridicality*). Likewise, on Boghossian's view of inference, bad inferences are inferences minus some property (e.g., *warrant transferring*).

Contra Boghossian, Quilty-Dunn and Mandelbaum (2018, 2019) tell us that *taking* is explicit only in "rich inferential transitions", and that inference proper casts a wider, and less intellectualized, net within our cognitive economy. Inferences in their base form—what Quilty-Dunn and Mandelbaum call *bare inferential transitions*, or "BITs"—need not be conscious and need not be "slow" or "reflective" in a dual-systems manner. A transition from mental state M1 to mental state M2 is inferential just when:

> (i) [M1] and [M2] are discursive, (ii) some rule is built into the architecture such that [M1] satisfies its antecedent in virtue of [M1]'s constituent structure and [M2] satisfies its consequent in virtue of [M2]'s constituent structure (*modulo* logical constants), and (iii) there is no intervening factor responsible for the transition from [M1] to [M2]. (Quilty-Dunn & Mandelbaum, 2018, p. 539).

We will return to the *discursivity condition* in (i) shortly. For now, when it comes to I2, two points bear emphasizing.

First, the rules in BITs are formal rules. Here is an example from Quilty-Dunn and Mandelbaum (2019, p. 154). Suppose that:

F(X)
IF F(X), THEN G(X)
∴ G(X)

is a rule. Since "the chair is black" and "the couch is brown" have F(X) as their structure, they can, as Quilty-Dunn and Mandelbaum note, be employed in inferences governed by this rule. And that these representations satisfy the structural specifications of this rule is unrelated to the content of their non-logical components. The rules only specify structural properties.

Second, a rule's being "built into an architecture" is not an appeal to dispositions.

It is, as Quilty-Dunn and Mandelbaum (2018, p. 540) stress, a counterfactual notion:

> In a world where there are no performance errors, the rule will accurately describe every transition within its scope. A mind can have such a rule built into it, even if the rule accurately describes only a small percentage of the transitions that mind is disposed to make in the actual world, due to systematic performance errors. A system can be disposed to make transitions in line with a rule without having that rule built into its architecture, and a system can have the rule built into its architecture without making transitions in line with it with any statistical regularity (Quilty-Dunn & Mandelbaum, 2018, p. 540).

For Quilty-Dunn and Mandelbaum (2018, p. 544), the notion of a *rule being built into an architecture* takes on much of the descriptive function of taking. The subject can appreciate the rule without taking an obvert attitudinal stand on it, and the rule "guide inferences directly". Again, there are *also* rich transitions, but that requires not just that the rule be actually built into one's central cognitive system, but also a disposition to endorse the transition itself (Quilty-Dunn & Mandelbaum 2018, p. 540).

The BIT-based case for I2 starts by noting that the criteria for BITs comes with no restriction on whether the rules are any good. Having a rule and the *normative assessment* of that rule are distinct. Affirming the consequent is a rule:

G(X)
IF F(X), THEN G(X)
∴ F(X)

It is a rule in that it is content-invariant and formal. But it is a bad rule. We are not saying that such a rule is in fact built into any architecture; but it could be, and that is what matters for I2.

Another BIT-based take on I2 requires a slight modification to Irrational-Think. On this tack, "incorrect inferences" are instead what Quilty-Dunn and Mandelbaum call "misinferences". Misinferences are *performance errors* (perhaps set off by some non-psychological intervening factor) and thus not cases where systems act in accordance with a rule. But, for an action to qualify as a performance error, inference rules must already be built into the system.[10] While this does not get us I2, we get a nearby claim that is good enough: LLMs *can* make inferences. And if they can do that, they can think (see fn. 8).

Finally, Buckner's (2019) categorization-based theory of inference takes the "lowest bounds" of inference to consist, roughly, in similarity-based associative categorization judgments.[11] The motivation here stems from the perceived failures of *taking*-style and BIT-style views (or "classic inferentialism") to accommodate inference in nonlinguistic subjects (e.g., prelinguistic human infants and non-human animals). In a broadly empiricist vein, Buckner rejects a sharp divide between associations and inferences, with inferences instead being a "rarefied" form of association. Inference is a *similarity-based*, as opposed to *rule-based*, categorization judgment. Buckner (2019, p. 705) gives an example of a population of lions in South Africa's Selous Game Reserve. Although lions do not typically hunt giraffes, these specific lions have learned that if giraffes get stuck in a sandy riverbed, they can go in for the kill without risking being kicked. The lioness learns from their individual experiences to "*flexibly interpret* an ambiguous cue—that a particular configuration of giraffe appearances is deadly in one context but can be safely pursued as prey in another" (Buckner, 2019, p. 705) By continuously associating experiences of a certain type (e.g., ones where giraffes look like *this*), the lion then makes a similarity-based categorization judgment that their current experience is *of this type*, motivating action. On this picture, the *taking*-condition is "finessed" (Buckner, 2019, p. 712), insofar as these categorical judgments are still made

[10] Performance contrasts with, and can mislead about, a system's underlying *competence* (Chomsky, 1965). Roughly, performance is what a system *does*, whereas competency is what a system *knows*—performance is a behavioral expression or use of this knowledge. For discussion of this distinction as it relates to human versus machine (including LLM) comparisons, see Firestone (2020).

[11] Buckner (2019, p. 698) is technically focused on *practical* (selecting actions from beliefs and desires) as opposed to *theoretical* rational inference (forming new beliefs). We assume, along with Buckner himself, that a similar enough story can be told for rational theoretical inferences.

in an intensionally-sensitive manner.

Buckner's categorization-based view also supports I2. Consider again the lioness. We can imagine a storm from the night before covering the ground with a layer of sand that is *just short* of being thick enough to trap the giraffe's legs, making the lioness' decision to hunt the giraffe the next morning decidedly bad. Nonetheless, there is still a similarity-based categorization judgment: given that the lioness' perceptual states are not fine-grained enough to pick up on the difference in sand-thickness, they thus judge that their current experience is of the suitable type. The result is a swift kick in the head, but one preceded by a (bad) inference.

Here is where we are. Recall again R1 from the argument from rationality:

**R1**. If LLMs think, they are rational.

R1, again, is false. Bad inferences *are inferences*. This claim has linguistic support, and it's true across vastly different accounts of inference. And to make inferences is to think. In lieu of R1, a better conditional is Irrational-Think: if LLMs are irrational, they think.

This is not to say that the antecedent this conditional is true. We will now argue that it is not, and if LLMs think at all, they probably just a-think.

## 4. If LLMs Think, They (Probably) A-Think

If LLMs think, they have mental states. And the kind of thinking we are interested in is thinking about the world. Thus, we assume that LLMs' mental states, if they have them, represent the world. Every drop of this is controversial. We cannot defend it here.[12] That is why our thesis is a conditional: *if* LLMs think, they probably just a-think. Assuming LLMs have mental states that represent the world, and make transitions between these mental states, our question is: what kind of mental transitions do LLMs make?

[12] Goldstein and Levinstein (2024) argue that LLMs plausibly have mental representations on several theories of mental representation (e.g., informational theories, teleological theories, structural theories), although they think it is inconclusive whether these representations count as propositional attitudes familiar from folk psychology (e.g., beliefs, desires). See also Herrmann and Levinstein (2024). For evidence that LLMs have *world models*, see Li et al. (2022) and Yildirim and Paul (2024).

Our answer: purely associative ones. To emphasize, this is a claim about mental processes, not the *instantiation base* of those mental processes. There is a version of *associationism*—understood as a thesis which *only* concerns the latter—which is trivially true for any neural net, connectionist-based architecture like LLMs. It does not follow from this that associationism—understood as the thesis that all mental processes are purely associative—is true (Mandelbaum, 2016, fn. 5; Mandelbaum, 2022). So, our thesis is hardly trivial.

## 4.1 A(nother) Word on Methodology

But how to defend it? Here is one idea, inspired by work on the distribution question for consciousness. Just as we can *begin* with a theory of consciousness, validated in humans, take that theory "off the shelf", and apply it to the disputed cases (e.g., non-human animals at various degrees of phylogenetic distance from humans, or even non-biological systems like LLMs), we can also start with a theory of inference, validated in humans, and take that theory "off the shelf", and apply it to the disputed cases. What, for example, would a BIT-style theory say about inference in LLMs?[13]

Questions like this intersect with the project of inner interpretability. For example, if it turns out that LLMs do not harbor discursive representations at all—something we presume inner interpretability could in principle discover—then LLMs cannot perform BITs and therefore cannot i-think.[14] And evidence suggests that transformer networks do not use variable binding (Gröndahl et al., 2023) and have limited compositionality and systematicity (McCoy et al., 2024), hallmarks of discursive or language-of-thought ("LoT") representations. And Klein (forthcoming) argues that, while LLMs *represent* structure, they likely do not make use of structured representations (due to permutation invariance). Without structured representations, there is no LoTs, and again, no i-thinking, given a BIT view.[15]

[13] The inspiration here stems from what Birch (2022) calls the "theory-heavy" approach to studying nonhuman consciousness. Birch rejects the theory-heavy view. Whether its (purported) problems generalize from studying consciousness to studying inference, however, we set aside.

[14] For another view of the format-inference connection, see Camp (2007).

That said, even if LLMs *do* harbor discursive representations, even those with full propositional structure, this is not sufficient for i-thinking on a BIT view. I-thinking, understood as BITs, requires that the transitions between representations result in virtue of their structure. In principle, one could have associative transitions between propositional structures; it is just that their logico-syntactic properties would be inert in effecting that transition (Mandelbaum, 2022).[16] What matters is the transition, not so much what the transition is between.

We thus take up our positive thesis via a different, behavior-based approach, one that operationalizes the i-thinking versus a-thinking distinction. We turn to that next.

## 4.2 Operationalizing Associative Transitions

The core starting idea is that association is, in some sense, *dumb*. This is not because it

[15] There are complicated, big-picture issues lurking in the background which require working through the relationship between BITs, the LoT thesis, and connectionism—in other words, the architectural basis of LLMs. Consider Chalmers' (2023) distinction between the *representational* LoT thesis (or "r-LoT") and the *computational LoT* thesis (or "c-LoT"). r-LoT says that the mind harbors discursively formatted, structured representations with propositional content. The c-LOT thesis is r-LOT plus an additional claim: that thinking involves computations defined over these mental representations. For some, connectionism is only compatible with r-LoT, not c-LoT. Unlike classicists, these connectionists take on a *non-concatenative* view of constituency structure (Rescorla, 2024). Here, the structure of a structured representation (like a LoT symbol) is encoded in a *distributed* fashion. Computation is sub-symbolic, and the vehicles of representation are not the objects of computation (Chalmers, 1990; Smolensky, 1998). So, we have r-LoT, but not c-LoT. How does this fit with BITs, if BITs require c-LOT too?

We will not belabor this, since our positive thesis will not appeal to any specific theory of inference. But it seems like the friend of a BIT view of inference will want to take on an implementational version of connectionism (if any). Chalmers is right that there is a *conceptual* distinction between r-LOT and c-LOT, but it is not clear what use r-LOT is without c-LOT; after all, if LOTs are not computed over, it is not clear how they can *do* anything (Quilty-Dunn et al., 2023b, p. 72-73). So, a proponent of BITs will treat connectionism not as a replacement paradigm (so not eliminativist), but as an account of how rule-governed symbol manipulation may be instantiated in physical systems (cf. Fodor & Pylyshyn,1988).

[16] Although that sort of set-up—where propositional representations were not also used in some *other* kind of mental process—would be a tad odd. At least in biological systems, there is evidence that propositional (and discursive) representations show up *throughout* the mind, including perception (e.g., Quilty-Dunn et al., 2023).

is simple, but because it is not amenable to rational considerations:

> Say one has just eaten lutefisk and then vomited. The smell and taste of lutefisk will be associated with feeling nauseated, and no amount of telling one that they shouldn't be nauseated will be very effective. Say the lutefisk that made one vomit was covered in poison, so that we know that the lutefisk wasn't the root cause of the sickness. Having this knowledge won't dislodge the association. [A]ssociative structures are functionally defined as being fungible based on counterconditioning, extinction, and nothing else (Mandelbaum, 2022, 4.4).

Three points need unpacking here: (i) associative structures, (ii) that of being fungible—or as we will put it, being modulable by counterconditioning and extinction, and (iii) the bearing of (i) and (ii) on being amenable to rational considerations.

To say that a mental transition is purely associative is to say that it is mediated by stored associative structures, and nothing else. Associative structures are analyzed functionally: *ceteris paribus*, if the mental state M1 directly activates the mental state M2, without intervening formal rules or alternative mental processes, then M1\M2 forms an associative structure (Quilty-Dunn & Mandelbaum, 2019).[17] An associative transition between PARIS and FRANCE occurs in virtue of a pre-existing structural-associative link between these mental states. Associations are often symmetric, but need not be (e.g., MANGO is more often associated with FRUIT, than FRUIT is with MANGO). And standardly, associative transitions are not formally rule-based.[18] The association itself does not *mean* anything at all; it is just a link that causes two mental representations to activate together (Dacey, 2019, p. 1208).

Extinction is a form of associative learning which works on pre-existing associations. Extinction modulates associations by repeatedly presenting a conditioned stimulus without its pre-associated unconditioned stimulus. Over time, this weakens the association between the two, reducing related conditioned responses (Mandelbaum,

[17] We assume that association does not result in predication (Fodor, 2003)—associating UNSCRUPULOUS and LAWYER does not predicate *unscrupulousness* of lawyers.
[18] Though see Shea (2024, pp. 19, 79-80) for doubts about this dichotomy.

2022). If the sound of a ringing bell always accompanies food, subjects learn to associate them and automatically salivate when hearing bells. After a while, when presenting subjects with numerous instances of the bell *without* the accompanying food, the stimuli are dissociated. The salivation response becomes extinct (Dunsmoor et al., 2015).

Counterconditioning is similar but involves association accompanied by valence (Mandelbaum, 2022). Valence does not need to be understood in terms of *conscious* affective states—which is good, since we are not assuming that LLMs have conscious states. Rather, valence might just be the non-conceptual representation of value (e.g., Carruthers, 2023). Understood this way, there is compelling evidence of nonconscious valence, as seen in the literature on subliminal motivation and instrumental conditioning (e.g., Pessiglione et al., 2007, 2008; Winkielman et al., 2005). As for the functional profile of valence, encouragement or discouragement of behavior is central. Counterconditioning changes associations by tagging them with new, opposite valences (Gast & De Houwer, 2013). For example, we might (cruelly) countercondition a dog with a shock collar so that whenever it tries to leave the yard, it receives a negative stimulus—eventually never attempting to leave.

Here, then, is the operationalization we have been building to. Let M1 and M2 be mental representations, where a *transition* between M1 and M2 is symbolized as M1 ⇒ M2. Mental transitions are sufficient for thinking (be it a-thinking or i-thinking). Then:

> For any mental transition M1 ⇒ M2, if M1 ⇒ M2 is directly modulable by only extinction or counterconditioning, and not rational intervention, then M1 ⇒ M2 is associative (cf. Mandelbaum, 2016; Quilty-Dunn & Mandelbaum, 2019).

Modulating a mental transition is a matter of somehow *breaking* it: breaking, for instance, one's propensity to move to M2 when in M1. Interventions that *merely* break, though, are too coarse. Chop off the right parts of one's brain, and you will disrupt a mental transition or two. But that would tell us nothing about whether the underlying transition is associative or inferential. Hence the appeal to being *directly* modulable. We need interventions that are (somehow) directed at the transition itself and give us a window into their nature.

So, associations are indeed dumb, but only in that rational considerations cannot be used to modulate them. As we will see, there is not just one way to understand rational modulation, but our argument fortunately holds up across these differences.

Note that this operationalization is just a way of making the a-thinking versus i-thinking distinction tractable. It simply assumes that if the best interpretation of a direct modulation is that it is not rational, but an instance of extinction or counterconditioning, then that is good evidence that the underlying targeted transition is associative, not inferential. We also stress that there is no *redescription fallacy* here. We are not saying it is impossible for LLMs to implement some cognitive capacity because all they do is linear algebra operations or next-token prediction (Millière & Buckner, 2024). In fact, the very possibility we are considering is one in which LLMs *do not* just do linear algebra operations or next-token prediction; to say that LLMs "merely" a-think is not to downgrade them from mindedness to something less.[19]

None of this is to insist that the distinction between associative and non-associative processes is perfectly sharp, or that associative learning theory is limited to counterconditioning and extinction paradigms (e.g., Buckner, 2011, 2019; Dacey, 2019). This does not change the main point, though, if all actual means of direct modulation in LLMs are indeed forms of counterconditioning and extinction.

### 4.3 The Argument from Arationality

We argue that:

[19] "Associative" is occasionally used in opposition to "cognitive", which might make one wonder how our thesis says *anything* about LLM minds or their being able to think. We suspect though that this gets perilously close to a verbal dispute over "thinking" and "cognitive". And, in the human case, we certainly have no trouble describing associative processes as instances of thinking. When one "jumps" from unscrupulous to lawyer, we describe this jump as thinking. One would say, "That made me think of lawyers!". In doing so, we are describing how tokening UNSCRUPULOUS led to a tokening of LAWYER, forming an associative structure of the form UNSCRUPULOUS\LAWYER. If associative transitions count as thinking in humans, why should they not count as thinking in LLMs?

A1. If a transition between M1 and M2 is directly modulable only via extinction or counterconditioning, but not by rational intervention, then, probably, that transition is associative.

A2. LLMs' mental transitions are directly modulable only via extinction or counterconditioning, and not rational intervention.

∴ LLMs' mental transitions are probably associative.

Since we are assuming that LLMs undergo mental transitions, our conclusion should be read as a conditional: *if* LLMs make mental transitions, these transitions are associative. Thus, *if* LLMs think, they probably just a-think. A1 falls out of our operationalization. Here we focus on A2. Whether A2 is true hinges on how we in fact modulate LLMs.

Given a pre-trained LLM, here are the candidate methods for modulating its behavior:

(a) Continued Pre-Training
(b) Fine-tuning
(c) Manually Adjusting Hyperparameters
(d) Chatting

Our position is this. Options (a) and (b) are forms of extinction or counterconditioning. Option (c) is not a direct method of modulation. And we urge that (d), while the especially tricky option to classify, is not *rational* modulation. Putting all that together, we have good evidence that LLMs' mental transitions are associative.[20] We will now look at each option in turn.

(a) Continued Pre-Training.[21] Roughly, during pre-training, an LLM makes a "guess" about the next token to appear in a provided sequence. The model's prediction is compared to the actual next token in the sequence and the difference between the two is measured as *loss*. A procedure called *backpropagation* computes gradients for each parameter, determining which parameters were most responsible for the loss. An optimization algorithm, such as stochastic gradient descent, then uses those gradients to update the parameters, reducing loss in future training rounds. This repeats in a cycle of prediction, loss calculation, backpropagation, and parameter-updating, which gradually improves the model's ability to predict tokens based on the training data.

[20] See Wang et al. (2024). Other emerging methods, such as machine unlearning and model editing, are set aside for the sake of space. We suspect that these techniques can be analyzed in a similar fashion.
[21] This is "continued" pre-training since we have an *already-existing* LLM (Parmar et al., 2024).

To illustrate, imagine a language model that has been trained solely on the sentence: "colorless green ideas sleep furiously". Given sufficient pre-training, the model accurately outputs "furiously" when prompted with "sleep". Take this to be the behavior we want to modulate. How can we do so by continuing pre-training? Say that we expand the domain of training data to the entire Wikipedia corpus. In this dataset, "sleep" is rarely followed by "furiously". During the continued pre-training process, the LLM's parameters slowly update to no longer assign high probability to "furiously" following "sleep". Given enough continued pre-training, when prompted with "sleep", the LLM outputs "peacefully" rather than "furiously". What sort of explanation ought we give to this process? We think this is clearly an instance of extinction. Recall how extinction works: the subject initially associates M1 with M2, and exhibits some related behavior B. Given enough examples of M1 without the accompaniment of M2, the subject slowly disassociates the two, and ceases to exhibit B. Almost identically, our toy LLM initially associates SLEEP with FURIOUSLY, and outputs "furiously" when appropriately prompted. After being presented with enough examples of "sleep" without "furiously", the LLM no longer associates the two, and its behavior changes.

(b) Fine-tuning. The aim of fine-tuning is to take an LLM pre-trained on a huge dataset of heterogeneous text and tune it to a specialized task. For example, ChatGPT was fined-tuned for dialogue, whereas the model from which it was tuned simply predicts next tokens. Here we focus on one powerful form of fine-tuning, *reinforcement learning by human feedback* ("RLHF"). Abstractly, a similar kind of process takes place in fine-tuning as in pre-training: the model makes a prediction, this prediction is scored, an algorithm is used to determine which parameters are most responsible for the score of the prediction, and those parameters are tweaked to improve the score on the next prediction. But in practice, there are differences between pre-training and fine-tuning. Specifically, during RLHF human evaluators give a score to the model's outputs, which signals whether outputs meet their preferences. Here, as opposed to pre-training, the model attempts to maximize its reward, rather than minimize its prediction loss. Often, when fine-tuning is conducted with RLHF, a *reward model* is used to score the target model's outputs. The reward model is trained on evaluated human responses and stands as a proxy for direct human feedback for the specified model. Either way, by the end of RLHF, the model's parameters have been adjusted such that it outputs text which better approximates the outputs human evaluators prefer.

Because of its seemingly valenced nature, we think this process is best understood as a form of counterconditioning. To elaborate, imagine a (pre-trained) model LLM which responds to users with phrases like "that's a stupid question". It is easy to imagine how this sort of response might do well to minimize prediction loss based on certain training data domains (like internet forums). Regardless, this is not behavior we want to see in our LLMs. So, during fine-tuning, human evaluators give low ratings to LLM outputs like "that's a stupid question". To maximize its evaluative scores, the model eventually ceases to output such phrases. Plausibly, what occurs here is that the negative score attaches an opposite valence to the existing association, consequently dissuading it from making the mental transition which leads to that output. Remember that valence here is understood functionally—a relationship between some discouragement and some behavior. When a negative score is given to an LLM's output, its parameters update to maximize expected score per output. Valence in biological systems plays the same role: a negative or positive valence in response to a behavior (especially if different than expected) induces an update of the expected value of that behavior, which in turn further encourages or discourages future activity (Carruthers, 2023).

(c) Manually Adjusting Hyperparameters. "Hyperparameters" is used in various ways. It can refer to the settings adjusted during the training process, like the learning rate or the number of epochs (rounds) the process should iterate through. In other setting, hyperparameters are settings adjustable at the time of inference, including settings like temperature and max tokens.[22] Here we will be using "hyperparameters" to refer specifically to the latter kinds of settings, since we take the former—as they are generally involved in training—to be covered in our pre-training discussion.

How do hyperparameters modulate LLM behavior? Let us focus on a model's *temperature*. At inference time, an LLM calculates the probability distribution of possible next tokens. The temperature hyperparameter determines whether the model outputs the single most likely next token, or instead samples more randomly. At temperature zero, an LLM deterministically outputs the unique most-likely next token. At higher temperatures, it rescales the logits in the softmax equation, giving tokens with lower likelihood a better chance at selection. This is seen as a way of controlling the determinacy of the model; at temperature zero, given the same prompt, an LLM will always output the same response. At higher temperatures, the model's responses become more unpredictable. For example, by dialing up the temperature in a model that always outputs "France" when prompted with "Where is Paris?" it will give more varied responses (e.g., "In France" or "France, which is in Europe").

Hyperparameter adjustment does not fit the mold of extinction or counterconditioning. Neither, though, is it a case of providing the LLM with something like rational considerations in favor of changing behavior. Instead, we will urge that the manual adjustment of hyperparameters fails to *directly* modulate the transitions themselves, being akin to physically manipulating the structure of a neural network. By comparison, in the cases of pre-training and fine-tuning, there is a direct relationship between the modulation process and the specific transition being affected. In fine-tuning, for example, an LLM outputs "That's a stupid question", and *in virtue of making this output*, gets a negative score from the reward model. As for the manual adjustment of hyperparameters, there is no direct connection between the specific target mental transition (e.g. WHERE IS PARIS? to PARIS IS IN FRANCE) and the method of intervention. So, while this is a way of modulating LLM behavior, it fails to do so directly.

[22] Here we mean the technical sense of "time of inference" (see fn. 1).

Imagine we attach an electrode array to your brain so that thoughts of food enter your mind whenever you try to solve math equations. Any reputable theory allows in-principle for physical interventions in the brain to manifest such mental effects. But this experiment does not tell us anything about your mental transitions prior to the physical change. That is, we expect such interventions to equally modulate inferential *and* associative transitions. The adjustment of hyperparameters is much like this. So, hyperparameter adjustment is not a direct modulation of LLM mental transitions at all.

(d) Chatting. Chatting with LLMs via conversational interfaces like ChatGPT is the primary way we interact with them. So, it is natural to wonder whether chatting is the best candidate for directly modulating LLM behavior in a way which is neither extinction nor counterconditioning, but rather something like rational modulation. After all, we can feed a model what seems like rational considerations and watch it respond appropriately. If there is evidence that not all LLM thinking is a-thinking, we suspect that chatting is the best candidate.

To make the concern especially sharp, consider *in-context learning* ("ICL"). Roughly, ICL is an LLM's ability to pick up on (novel) patterns or instructions in inputs. In standard tests of ICL, an LLM is given a prompt of input/output pairs, and a request to make a prediction about the final input/output pair (Akyürek et al., 2022). Here's a simplified example:

*Prompt*:

x=1 → y=3

x=2 → y=5

x=3 → y=7

x=4 → ?

Models notice the pattern in the prompt, "see" that the relationship is linear, and (by hypothesis) approximate the linear regression algorithm to map x's to y's, using the determined linear relationship (e.g., $y = 2x + 1$) to make an accurate prediction.

*Output*:

y=9

There are, unsurprisingly, many questions about how best to interpret ICL, and some answers are not obviously friendly to i-thinking and rational modulation (Dong et al., 2024). But we grant that it looks like ICL really is learning, and we grant that it does not look terribly like associative learning, *especially* if ICL involves the deployment of rules.

Yet ultimately, ICL—and chatting more generally—doesn't show that LLM's engage in i-thinking. One initial cause for skepticism here is that, during chatting, an LLM's weights are frozen (Millière & Buckner, 2024). So, from the outset it is a bit forced to speak of *modulating* LLMs in any way during chatting.

That the weights are frozen is indeed important. But to get at the crux of the issue we should first step back and take a closer look at what a rational modulation in an LLM might be. Once that issue is pinned down, we can *then* ask whether chatting fits the conditions. To that end, we will consider two ideas. One will require that rational modulation consist (in part) in the engagement of motivating reasons; the other will require only a certain kind of attention shift. Neither will cut it.

Inference is itself *reason-responsive* (Quilty-Dunn & Mandelbaum, 2019, p. 160; Mandelbaum, 2016, p. 636). So, the first idea is to say that, for a modulation to be evidence of inference, it must provide a *reason*, and it must affect a *response* to that reason—specifically, a *motivating reason*, not (just) a *normative one*. This latter distinction is familiar: whereas a normative reason is simply something that counts in favor of S's ϕ'ing, a motivating reason is one that in fact guides ones' actions. These can come apart: that *p* is a reason to ϕ says nothing, on its own, about why S ϕ'ed. (Being healthy is a reason for eating vegetables, but it may not be why S eats vegetables; maybe S eats vegetables to placate her parents.[23]) The problem is that, on the main accounts of motivating reasons, chatting inputs—or the information encoded by them—*simply cannot constitute motivating reasons*. If so, they cannot be employed as part of a rational modulation of an LLM.

[23] Motivating reasons are often explanatory, but they need not be, as when what motivates S is a false belief (Alvarez, 2008). However, since in the present case we are interested in *explaining why* a transition M1 ⇒ M2 has been modulated, we focus on motivating reasons that explain. And since we care about *rational* modulation, we do not want to rest with mere *causal* reasons.

Let us grant that chatting inputs are (at least) normative reasons. This speaks to the initial intuitive pull behind chatting as a means of rational modulation: what is communicated by a chatting input such as, "Baltimore is not in Texas" certainly provides a reason for no longer thinking that Baltimore is the capital of Texas. Call this specific input *p*. Now, is it also a motivating reason? Consider *psychologism* about motivating reasons. This view, from Davidson (1963), says that the motivating reason is not *p*, but a specific psychological state, viz. S's *belief* that *p*. However, given psychologism, chatting cannot serve up motivating reasons. As standing states, beliefs are plausibly encoded in an LLMs' weights (Chalmers, 2025). Yet as noted, during chatting, the LLMs' weights are frozen; indeed, part of the marvel of ICL is an LLM's ability to seemingly learn novel tasks without any change to its weights (Akyürek et al., 2022).

Note that the claim here is *not* that LLMs do not have beliefs. Maybe they do; maybe they do not. (Our claim about a-thinking notwithstanding.) The question, at this stage, is simply whether chatting is really a form of rational modulation, and so evidence of i-thinking. So, what we are claiming here is not that LLMs do not have beliefs *simpliciter*, but that chatting inputs cannot be encoded as beliefs, since acquiring a belief would require a change in the LLM's weights. Therefore, given psychologism, chatting inputs cannot be motivating reasons.

Suppose though that instead of psychologism, we opt for *propositionalism* (Alvarez, 2008). Here a motivating reason is not my *belief* that p, but simply the proposition *p*. A similar problem emerges, however. For unless the subject *believes* that *p*, the reason—here, just *p*—cannot be *her* reason, nor can *she act for* that reason (Alvarez, 2008, p. 56). Beliefs are thus inescapable. Again: chatting inputs cannot be taken in as a belief, because an LLMs' beliefs (such as it has any) will supervene on its weights, and chatting does not alter an LLMs' weights.[24]

[24] Even on *factualism* (Dancy, 2000), where motivating reasons are not beliefs or propositions, but *facts*, the point still holds. In general, there must be some cognitive relation R such that S stands in R to the reason, in order for S to act *for that reason*. R is likely no weaker than belief (*knowledge* being stronger), and this is so no matter one's ontology of reasons. The only difference with factualism is that R takes S on an indirect route to the reason: R directly relates S to a proposition which is *about* that fact.

But maybe all of this is too demanding of an account of rational modulation. Maybe the engagement of motivating reasons, and so changes to beliefs, and so changes to weights, are not necessary. So, here is a different idea. Assume for the moment that some forms of *attention* can be rationally appraisable (e.g., Siegel, 2017). Then, notice that during chatting an LLM's attention can be directed at more or less salient features of the input. The idea would then be that attention is the mechanism by which chatting *rationally* modulates LLM mental transitions.

It is not obvious that *attention* is precisely what is picked out by "attention" in the context of transformer-based architectures.[25] But waving this, there is a more pressing issue: in virtue of what, exactly, is attention rationally appraisable? We clearly do not think of *every* instance of attention as rationally appraisable; bottom-up attention—like snapping towards an unexpected gunshot—does not seem to be. Siegel's (2017) answer to this question reveals the problem: attention is rationally appraisable when, and only when, it is controlled by the conclusion of an inference. To use Siegel's (2017, p. 178) example, when examining CVs your decision to attend to certain features to the exclusion of others is irrational when it is based on a conclusion one *inferred* from a prejudiced belief. This is of no help. For now, to determine whether the attention mechanism in LLMs is the rationally relevant kind of attention, we would have to know whether LLMs were performing an inference—we would have to know if they were i-thinking. So, we are right back at our original question.

## 5. Conclusion

To sum up: we have made a negative and positive point. The negative: Stoljar and Zhang's argument from rationality fails. It is not true that if LLMs think, they are rational. For if LLMs are irrational, they *do* think. They *i-think*. (Indeed, even an LLM's being *arational* is consistent with their thinking.) But we are doubtful that LLMs *really* i-think. Hence our positive claim: if LLMs think, they probably (just) a-think. Continued pre-training and fine-tuning are direct means of modulating LLMs, but they are best construed as forms of extinction and counterconditioning. Hyperparameter adjustment is only an indirect way of modulating LLMs. And while chatting, supposing it is even a form of modulation at all, does not seem to be an instance of extinction or counterconditioning, it is not a form of rational modulation either. Or our being able to tell whether it is requires us to *presuppose* that LLMs are already capable of inference. At the very best, then, chatting does not improve our evidential situation with respect to the kind of minds LLMs may have. Thus, on the most reasonable interpretation of LLM modulations, if LLMs have minds, they are likely *purely associative minds*.

[25] Roughly, an "attention block" computes the weighted similarity of the key and query values between each token and every other token in the input, producing an attention matrix. This matrix encodes each token's relevance to every other token, which is then used to update those tokens' embeddings (Kim et al., 2017). This is what allows transformers to outperform recurrent neural networks, as it vastly improves the context-awareness (e.g., discerning which sense of "bank" is relevant) of the language model.

**Acknowledgements**

We would like to thank the editors of *Mind and Language*, along with the two anonymous referees for helpful feedback. Further thanks to the following for helpful feedback and discussion on earlier drafts of this essay: Jacob Berger, Cameron Buckner, David Chalmers, William D'Allessandro, Ali Rezaei, Daniel Stoljar and Zhihe Vincent Zhang.